\documentclass[10pt, conference, compsocconf]{IEEEtran}

\usepackage{mathtools}
\usepackage{graphicx}
\usepackage{amsmath}
\usepackage{booktabs}
\usepackage{multirow}
\usepackage[dvipsnames]{xcolor}

\usepackage{color, colortbl}
\definecolor{LightCyan}{rgb}{0.88,1,1}

\usepackage{stmaryrd}

\usepackage{wrapfig}
\usepackage[normalem]{ulem}
\usepackage{multicol}

\def\update#1{#1}

\usepackage[hidelinks]{hyperref}
\hypersetup{
    colorlinks = false,
    linkbordercolor = red,
    citebordercolor = green,
    urlbordercolor = blue,
    pdfborder = {0.5 0.5 0.5}
}

\begin{document}

\title{NeuralPrefix: A Zero-shot Sensory Data Imputation Plugin}

\author{\IEEEauthorblockN{ Abdelwahed Khamis}
\IEEEauthorblockA{\textit{Data61}\\
\textit{CSIRO}\\
Brisbane, Australia \\
abdelwahed.khamis@data61.csiro.au}
\and
\IEEEauthorblockN{Sara Khalifa}
\IEEEauthorblockA{\textit{School of Information Systems} \\
\textit{Queensland University of Technology}\\
Brisbane, Australia \\
sara.khalifa@qut.edu.au}
}

\maketitle

\begin{abstract}
Real-world sensing challenges such as sensor failures, communication issues, and power constraints lead to data intermittency. An issue that is known to undermine the traditional classification task that assumes a continuous data stream. Previous works addressed this issue by designing bespoke solutions (i.e. task-specific and/or modality-specific imputation). These approaches, while effective for their intended purposes, had limitations in their applicability across different tasks and sensor modalities. This raises an important question: \textit{Can we build a task-agnostic imputation pipeline that is transferable to new sensors without requiring additional training?} In this work, we formalise the concept of zero-shot imputation and propose a novel approach that enables the adaptation of pre-trained models to handle data intermittency. This framework, named NeuralPrefix, is a generative neural component that precedes a task model during inference, filling in gaps caused by data intermittency. NeuralPrefix is built as a continuous dynamical system, where its internal state can be estimated at any point in time by solving an Ordinary Differential Equation (ODE). This approach allows for a more versatile and adaptable imputation method, overcoming the limitations of task-specific and modality-specific solutions. We conduct a comprehensive evaluation of NeuralPrefix on multiple sensory datasets, demonstrating its effectiveness across various domains. When tested on intermittent data with a high 50\% missing data rate, NeuralPreifx accurately recovers all the missing samples, achieving SSIM score between 0.93-0.96.  Zero-shot evaluations show that NeuralPrefix generalises well to unseen datasets, even when the measurements come from a different modality. Project page: \href{https://neuralprefix.github.io/}{https://neuralprefix.github.io/}

\end{abstract}

\begin{IEEEkeywords}
 Data Imputation, Modality Compensation, Neural ODE
\end{IEEEkeywords}
  
\maketitle

\section{Introduction}

Data intermittency is one of the most important problems encountered in the sensing and pervasive computing domains. Unavoidable issues such as transient failures, communication issues, and power constraints lead to intermittent observations that are known to negatively impact the data analysis pipelines and machine learning models acting on the observations. Such a critical problem has been addressed by many research works \cite{adhikari2022comprehensive}.  Most notable among the proposed techniques are the learned generative models \cite{gao2022generative} that made a remarkable progress in spatio-temporal data imputation. These models \cite{yuan2022stgan} are trained to fill the missing data points by conditioning on observed values.

\begin{figure}[th]
    \centering
    \includegraphics[width=0.95\linewidth]{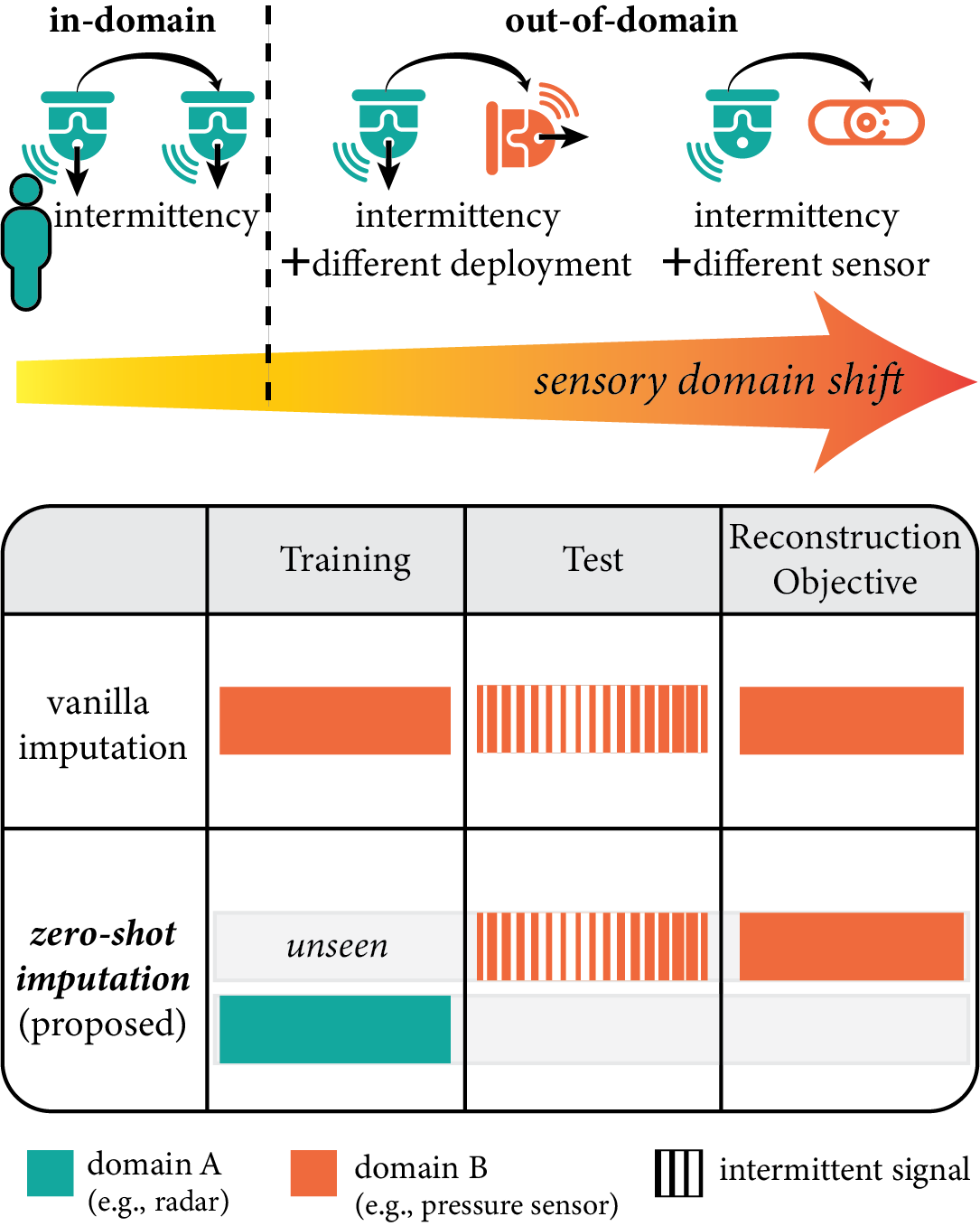}
    \caption{\textbf{Zero-shot Imputation (ZSI)}. We frame the new problem of zero-shot imputation and highlight potential applications (Fig. \ref{fig:application}). Given \update{sensory data} in a specific domain (A), ZSI seeks to learn an imputation model that can be transferred to unseen domains (B) without re-training. The table above contrasts the idea to vanilla imputation. \textit{Vanilla imputation} handles intermittency in \update{the} same domain (e.g. same sensor) . 
    \textit{Zero-shot imputation}, on the other hand, generalises imputation capabilities to domains (modalities) unseen during training.  
    }
 \label{fig:signature}
 \vspace{-1em}
\end{figure}

\begin{figure*}[t!]
\centering
  \includegraphics[width=0.9\textwidth]{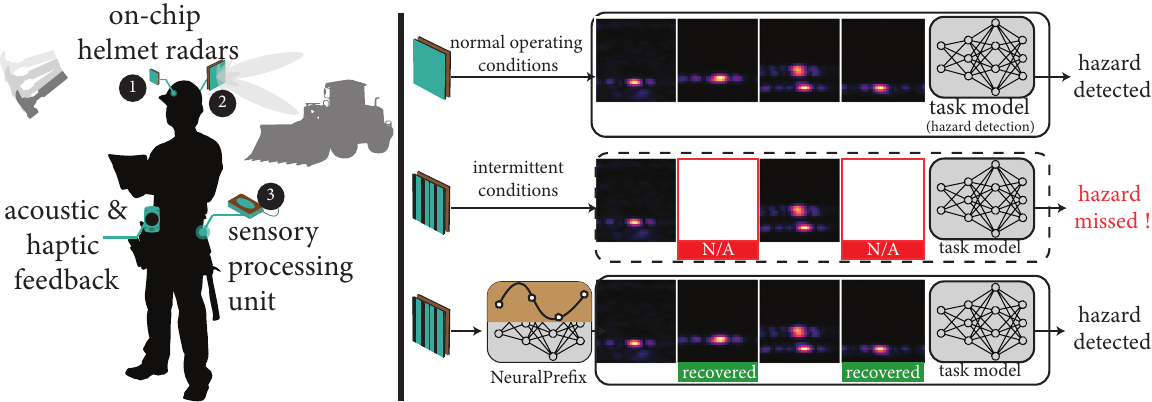}
  \caption{\textbf{Example application of Zero-shot Impuation}. (Left) A hazard perception suit encompasses several wearable sensors, such as on-chip radars, to sense the ambience and inform/warn the wearer of important events, including quickly falling objects or approaching vehicles. Intermittency (e.g. transient sensor failure) can result in missing those events and hence dangerous consequences.  (Right) NeuralPrefix can preface the existing model to reconstruct the complete signal \underline{in a zero-shot manner} (i.e. generalising to unseen modalities).  It doesn’t require modifying either the sensor's operational cycle (on/off schedule) or the underlying ML model.}
  \label{fig:application}
\end{figure*}

A common assumption in learned data imputation is that the test domain is homogeneous to the training domain. Thus, the training and testing setups are required to represent the same task (e.g. human walking action) and/or the same sensory modality (e.g. Radio Frequency (RF) Sensing). Overcoming this limitation would enable exciting opportunities in the sensing domain.  For example, a generic data intermittence handling mechanism can be developed and ``pooled'' among numerous sensors (e.g. radar and pressure sensor in Fig.~\ref{fig:signature}), allowing new sensors to use it without cumbersome re-training efforts.

\textbf{Zero-shot imputation (ZSI) (Figure. \ref{fig:signature}).} Inspired by the concept of zero-shot learning (ZSL) \cite {pourpanah2022review}, this paper introduces zero-shot sensory data imputation as a key enabler for handling missing data across diverse sensors without requiring re-training. ZSL is a paradigm where a model can make predictions on categories or tasks it has never seen during training by leveraging prior knowledge that relates to the unseen data. Given this, a question that emerges is \textit{Can we develop a learned data imputer that generalises to unseen sensory modalities without re-training?} This paper answers this question affirmatively.

\begin{figure}[t!]
    \centering
    \includegraphics[width = 1\linewidth]{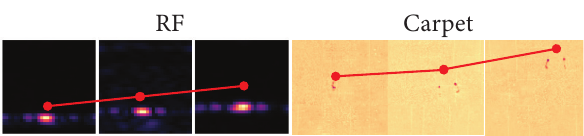}
    \caption{\textbf{Common dynamics exist in different datasets in the data space despite the modality disparity.}}
    \label{fig:common_dynamics}
\end{figure}

To achieve this, we can build on the observation that intermediate missed frames can be thought of a product of the interaction between appearance (content) and dynamics (motion) features. While appearance features vary across sensor modalities, underlying dynamics (human motion) tend to be similar across dataset and can therefore be leveraged as the zero-shot prior. To illustrate, consider the example sensory data in Figure.~\ref{fig:common_dynamics} from RF and pressure mat sensors: the left RF frames show a hand gesture, while the right frames depict walking on a pressure mat. Despite the disparity in actions and modalities, common dynamics can be observed in the data space. Specifically, the brightest blob in each frame follows a semi-linear trajectory (motion) while being continually deformed (content) over time.
This example reflects a broader trend in real datasets, where sensors capture natural phenomena that tend to change smoothly over space and time, following certain continuous physical dynamics.  By learning these dynamics, we can impute the missing signal at any time step. Therefore, we propose leveraging these smooth and predictable changes to design an effective \textit{generative model that weighs more on the dynamics side (rather than content)}. This approach enables sensory data imputation pipeline that can be \textit{transferred to novel sensory modalities in a zero-shot manner (without the need for re-training)}.

To build on this, we introduce a novel independent component, NeuralPrefix, which can preface the task model and recover the missing \update{spatio-temporal} data samples before they are consumed by the task model (see Fig. \ref{fig:application}).  Prior to deployment, the prefix model can be calibrated to learn the continuous dynamics of the target data. The calibration is done through a \textit{commensal training} \footnote{In biological interactions, a commensal organism benefits from attachment to host species without affecting the latter. Analogously, the prefix model nourishes on the unlabeled task training data (or a similar dataset) without affecting the task model. }
process. That is; training on an unlabeled version of the task dataset with the objective of sample reconstruction.

In this direction, we suggest integrating Neural Ordinary Differential Equations (Neural ODE) \cite{chen2018neural} into the generative component. This concept is fundamentally different from traditional networks (e.g. RNNs) whose internal states are defined at discrete time points. In contrast, Neural ODE models the hidden state as a continuous trajectory represented by ordinary differential equations. 
By using an ODE solver, the hidden state can be integrated from one observation to the next, regardless of the spacing between time points, making it practically suitable for intermittent observations. The solution is further improved by addressing the sparsity of sensory data frames. The shrinkage loss \cite{lu2020deep} is leveraged to focus the reconstruction budget only on the important and hard-to-construct parts (foreground pixels) while tolerating the background and noise.

This work makes the following contributions:

\begin{itemize}
    \item Formalising the concept of zero-shot sensory data imputation as a key enabler for handling missing data
across diverse sensors without requiring re-training.
    \item Introducing NeuralPrefix; a novel and simple architectural paradigm (\textit{prefix model}) that decouples the data intermittency handling mechanism from the target model. Thus, upgrading traditional models with intermittency robustness without re-training or fine-tuning. To the best of the authors' knowledge, this is the first task/sensor-agnostic data intermittency handling solution. 
    
   \item Comprehensive experimental demonstration of NeuralPrefix by evaluating it using large-scale spatio-temporal human action datasets, which include two RF datasets and one tactile carpet dataset, with extensive comparisons against several traditional and recent imputation approaches.
\end{itemize}

\section{Related Work}
\label{sec:related_work}

\update{\textbf{Data Imputation.}} The field of data imputation has seen a broad range of techniques, from traditional statistical methods to more advanced approaches. Early methods like K-Nearest Neighbors (KNN) \cite{knn} and probabilistic approaches such as Expectation-Maximisation (EM) \cite{EM} are simple to implement but often serve as baselines due to their inability to capture spatial and temporal dependencies, resulting in limited recovery quality for practical use. Spatio-temporal multi-view learning \cite{ST-MVL} improves data imputation by considering both spatial and temporal dependencies, approximating missing values using low-rank matrix completion \cite{LRMC,ensemble} and tensor decomposition \cite{tensor,tensor2}. However, these methods often fail to capture the deeper, inherent dependencies in spatio-temporal sensor data.

\textbf{Deep Generative Models.} Recently, deep learning approaches have emerged for data imputation \cite{wang2024deep, sun2023deep}, including the use of generative models \cite{adhikari2022comprehensive, shahbazian2023generative}, offering more advanced solutions to address these limitations.
This work falls in the category of generative data imputation by building on the same principles. To the best of the author's knowledge, this is the first work to employ generative modelling for zero-shot sensory data imputation. Beyond this main distinction, the framework possesses a number of technical qualities that collectively enable a broader range of applications. Specifically, the capability to perform interpolation and extrapolation using an independent parameter-efficient design, where the Neural ODE parameters and memory budget are fixed regardless of the sequence sizes. These qualities collectively enable a broader range of applications.

\textbf{Neural ODE works.} Our architecture is inspired by the Neural ODE-based video generation models, including MODE-GAN  \cite{kim2021continuous} , VidODE \cite{park2021vid}, and \cite{kim2021continuous}, with a key distinction in its focus on out-of-domain generalisation; an aspect not explored in these studies. Additionally, NeuralPrefix is designed for sparse (\update{lower-dimensional}) sensory data rather than vision data. For this, our approach is purposefully simpler (e.g. no GAN as followed in \cite{park2021vid, kim2021continuous}), and the design choices tailored for the considered data type, such as the shrinkage loss (Sec.~\ref{sec:architecture}). 

\update{\textbf{Time Series Foundational Models.} Concurrent with our efforts, recent advances in foundational models show the potential of task-agnostic data imputation (among other tasks).  Inspired by Large Language Models (LLMs), these works \cite{ansari2024chronos}  posit that a time series model trained on a large collection of multi-domain datasets can generalise to unseen domains without re-training (i.e. zero-shot). While being a promising direction, these models are extremely large, making them much more costly to train and less practical for sensory applications.  For example, Amazon's Chronos \cite{ansari2024chronos} leverages the T5 Transformer (710M parameters) for time series, requiring 504 hours of training on A100 GPU, whereas NeuralPrefix takes a mere 6 hours of training on RTX4090. }

\section{Method}

\textbf{Brief Notation}. Lowercase bold symbols ($\mathbf{v}$) denote vectors and uppercase bold symbols ($\mathbf{V}$) for tensors. The symbol $\big\{i\big\}_{1}^{N}$  denotes the sequence
$[1, \cdots, N]$.  

\textbf{Problem.} Our goal is developing the prefix  
$\mathcal{G}$
that takes an intermittent input sequence $\mathbf{X}$ whose data samples \update{are observed} at points $\{t_i\}$ and recovers the data points impacted by intermittency at the temporal coordinates $\{m_i\}$. 
Specifically, the prefix model is trained on the intermittence-free dataset similar to that used for training the target model. 
This simple design, given that the prefix model is \update{detached from the host model} and target task, is very flexible. \update{Thus,} one can use the same prefix for multiple sensors (task models) or even multiple modalities (as we will see in Sec.~\ref{sec:out_of_domain}) 

Formally,  the problem is framed as sequence to sequence regression. Specifically, we can build a generative model to perform the mapping
$\mathcal{G} : \big\{\mathbf{x}(t_i)\big \}^N_{i=1} \mapsto \big\{\mathbf{x}^{\prime}(m_i)\big\}^K_{i=1}$. 
The input data frames $\mathbf{x}(t_i) \in \mathcal{R}^{W\times H \times C}$
, where  $W, H$ and $C$  denote width, height and number of channels; respectively, aren't uniformly spaced in time
(i.e. $(t_i - t_{i-1}) \neq (t_{i-1} - t_{i-2}) $). 
Formulated this way, this mapping is interpolation if we set $m_1 > t_1$ and $m_K<t_N$ and extrapolation if  $m_K> m_1 > t_N$. 

At \update{a high} level, we build $\mathcal{G}$ as Encoder-Decoder architecture (Sec. \ref{sec:architecture}) with a learning objective of minimising the data reconstruction error. During training, we simulate data intermittency by masking some samples from the input sequences and tasking the model with reconstructing them.  
Yet, modelling the sporadic data is challenging, and the key idea is to treat the \update{latent state} as a continuous trajectory. \update{The next} section introduces this idea.

\subsection{Continuous Latent Dynamics Primer}
\label{sec:latent_dynamics}

\begin{figure}[t]
    \centering
    \includegraphics[width=1\linewidth]{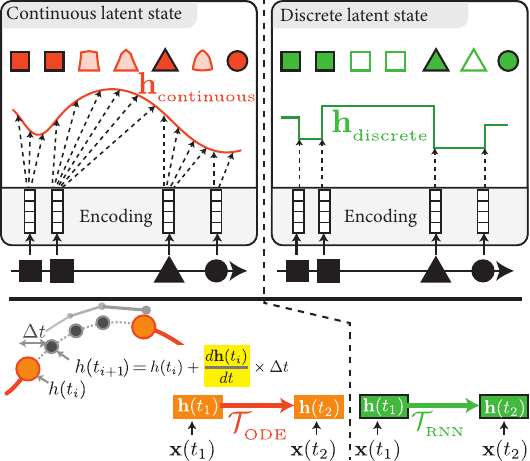}
    \caption{\textbf{Continuous vs Discrete Latent States in Neural Networks.} (Bottom Left) state estimation using a form of ODE. The state at time $t_i$, denoted as $\mathbf{h}(t_i)$, is updated to next state $\mathbf{h}(t_{i+1})$ through the equation above. Since intermediate steps ($\Delta t$) can be arbitrarily small, the continuous trajectory (curved path) can be obtained. }
    \label{fig:continuous_latent_state}
\end{figure}

A typical data-driven approach for modelling sequence-to-sequence prediction is recurrent modelling (e.g., RNN). The input data is processed sequentially with the help of internal state $\mathbf{h}(t_i)$ corresponding to the temporal coordinate $t_i$ as follows:
\begin{equation}
 \mathbf{h}(t_{i}) = \text{RNN}(\mathbf{h}^{\prime}(t_i),  \mathbf{x}(t_i)) \ , \ \mathbf{h}(t_{i-1}) \xrightarrow{\mathcal{T}_{{}_\text{RNN}}}  \mathbf{h}^{\prime}(t_i) 
 \label{eg:rnn}
\end{equation}
where $\mathcal{T}_{{}_\text{RNN}}$ manages the state transition across temporal steps in a uniform manner regardless of the temporal gap between incoming samples. Since Equation \eqref{eg:rnn} assumes uniformly spaced data samples, its internal state is kept unchanged until the next data point arrives
as shown in Fig.~\ref{fig:continuous_latent_state} (\update{top} right). Consider the long temporal gap between the square and triangle in the figure. The internal state (and thus the output) won't evolve in that stretch due to the discrete nature of $\mathcal{T}_{{}_\text{RNN}}$. 
A possible solution is to gradually deviate from the last state as time progresses using approaches such as temporal decay \cite{rajkomar2018scalable}. A better alternative is reflecting the continuous nature of the incoming signal into the model's latent space by requiring the internal state transitions to follow a continuous trajectory (identified by an ODE). In effect, combining the data with an ODE state transition prior $\mathcal{T}_{{}_\text{ODE}}$. Thus, at any temporal point, even when the observation is missing,  the model will estimate the internal state trajectory by relying on the last known state 
and the expected state dynamics (i.e. the rate of change) as depicted in Fig.~\ref{fig:continuous_latent_state} (bottom left). In other words, we frame the problem as state estimation in a continuous dynamical system.

Specifically, and given $\mathbf{X}$,  we seek to compute the state $\mathbf{h}(t) \in \mathcal{R}^{d}$ at arbitrary times $t_i$ (equivalently at intermittent points $m_i$).  We compensate for missing data by relying on a component $\mathbf{g}(\mathbf{h}(t), t) = \frac{d\mathbf{h}(t)}{dt}$ that captures the evolution of model's internal state. 
Initially, assume that $\mathbf{g}(\mathbf{h}(t), t)$ is given (we relax this later).
Given the above, $ \mathbf{h}(t)$ can be determined from the last known state at $t_k$, where $t_k<t_i$,  as follows:
\begin{equation}
\mathbf{h}(t_i) = \mathbf{h}(t_k) + \int_{t_k}^{t_i}  \mathbf{g}(\mathbf{h}(t), t) dt
\label{eq:ivp}
\end{equation}

Intuitively, Equation \eqref{eq:ivp} is an ODE initial value problem (IVP) that continually integrates the rate of change (of the system's state) to find the next state. One step of this process (in a \update{discrete-time} step) is depicted in Fig.~\ref{fig:continuous_latent_state} (top). This is fundamental in solving problems involving dynamic systems. 

Before continuing,  we note two things. First, the dynamics of the hidden state $\mathbf{g}(\mathbf{h}(t), t) $ aren't known in practice. We follow \cite{chen2018neural} and parametrise that component as a neural network $\mathbf{g}_{\theta}(\mathbf{h}(t), t)$, where $\theta$ denotes the network's weights, to learn the dynamics in end-to-end manner. Second, there is no \update{closed form} solution for the integral in Equation \eqref{eq:ivp}. However, off-the-shelf ODE solvers, such as Euler and Runge-Kutta, can be leveraged. Consequently and considering the changes, we can re-write Equation \eqref{eq:ivp} as :
\begin{equation}
    \mathbf{h}(t_{i+1}) = \overbrace{\textbf{\text{ODESolve}}}^{\text{black box ode solver}}(\underbrace{\mathbf{g}_{\theta}}_{\text{dynamics}},\underbrace{\mathbf{h}(t_{i})}_\text{initial state},\ \underbrace{(t_{i}, t_{i+1})}_{\text{time steps}}\ )
\label{eq:ode_solve}
\end{equation}
where \textbf{\text{ODESolve}} abstracts the integral computation within a black-box ODE solver, which offers interesting benefits. The solver can be upgraded/adapted during the inference time without requiring re-training. It allows for different trade-offs between computational efficiency and accuracy at various stages of the model's deployment. Note that the end-to-end network training can be done efficiently in the presence of a black-box solver using the Adjoint Sensitivity Method. Details were omitted due to space limitations. The interested reader can consult \cite{chen2018neural}. 

\begin{figure}[t!]
    \centering
    \includegraphics[width= 1\linewidth]{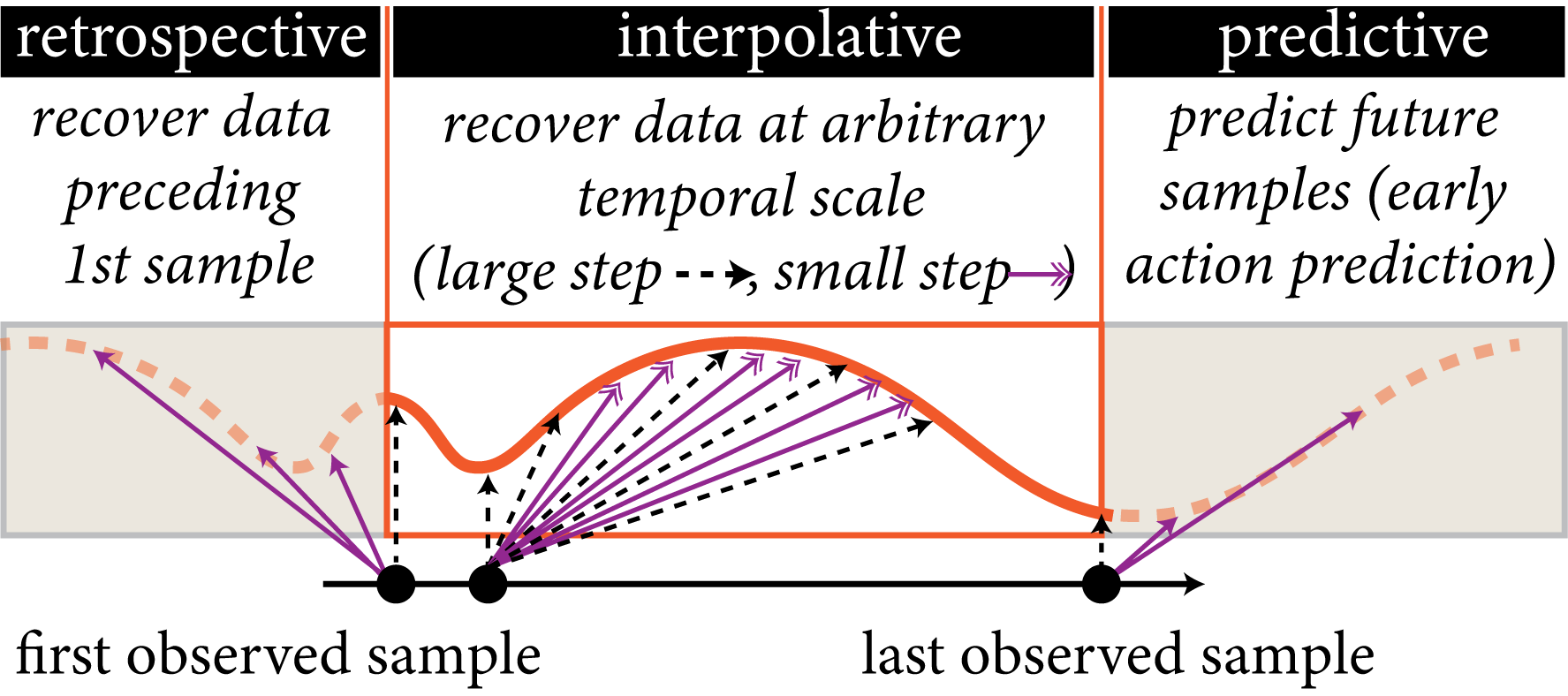}
    \caption{\update{\textbf{ODE data recovery modes include interpolative, predictive and retrospective. }}}
    \label{fig:forward_backward_mode}
\end{figure}

A unique property of \update{continuous modelling is the flexibility} that allows catering for various intermittency modes and applications, \update{as shown in Figure~\ref{fig:forward_backward_mode}}. In addition to interpolation, the prefix can be configured on extrapolation. Thus, upgrading the task model with early action prediction \cite{wang2019progressive,kong2017deep} capabilities. Even more interesting, as ODE can be solved in the backward direction, retrospective discovery can be done. Thus, one can estimate the dynamics preceding the first incoming frame. 

So far, we demonstrated the mechanism by which we can enable continuous state transition in a neural network. Next, we explain how this mechanism is used in NeuralPrefix encoder-decoder architecture.

\begin{figure*}[t]
    \centering
    \includegraphics[width=1\linewidth]{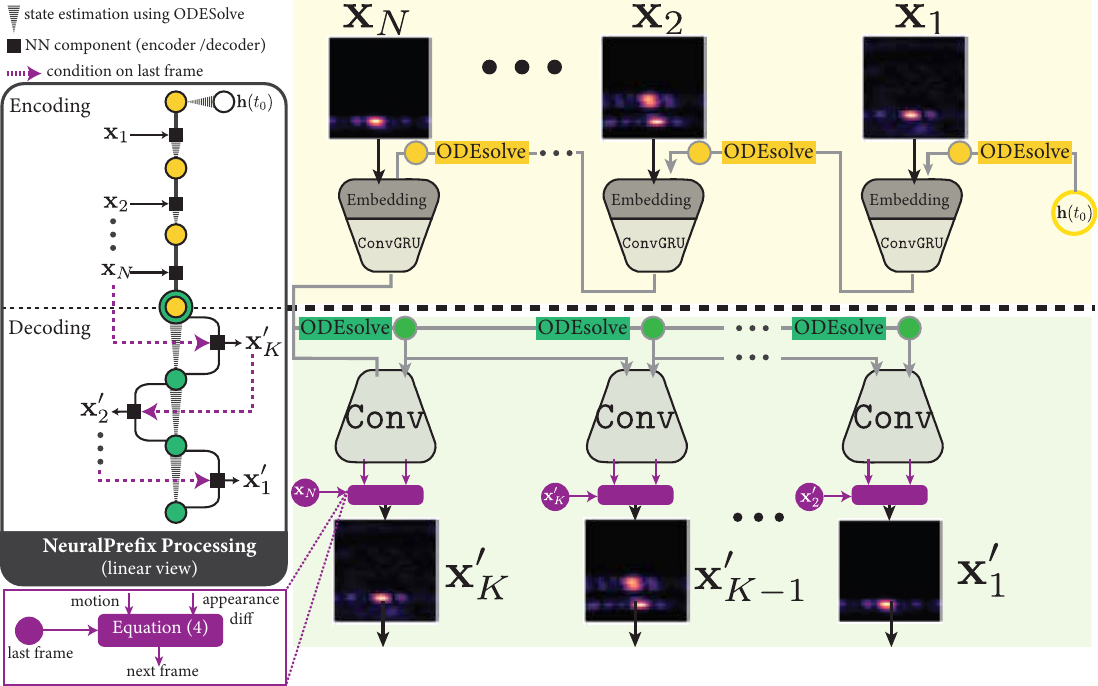}
    \caption{\textbf{NeuralPrefix Plugin.}
    NeuralPrefix is a continuous Encoder-Decoder architecture that imputes data in an autoregressive manner. Internal (latent) state transitions are governed by Ordinary  Differential Equations. (Left) simplified linear view \update{of the} architecture. (Right) unrolled view of the architecture showing the progression of data generation process. 
    Modular Frame Generation (purple box) uses Eq.~\ref{eq:composition} to compose the decoder's outputs into the target frame.}
    \label{fig:arhcitecture}
\end{figure*}

\update{Before presenting NeuralPrefix architecture, we note a key distinction between the dynamics in NeuralODE  ($\mathbf{g}_\theta$) and those in traditional (neural network-free) differential equations ($\mathbf{g}$). Such distinction makes each suitable for a different scenario depending on the application's requirements.}

\update{\textbf{Interpretability and Data-Efficiency.} Indeed, \textit{standard differential equations} can be used for imputation applications.  
For example, in a wearable sensing context, one might model human arm movements as a damped harmonic motion.} \update{Thus, the dynamics $\mathbf{g}$ are explicitly captured (not learned) using the differential equation $\ddot{x} + 2\zeta\omega_0\dot{x} + \omega_0^2 x = 0
$ where $x$ is the displacement from the equilibrium (rest position), $\omega_0$ is the natural frequency of the system and $\zeta$ is the damping ratio. After calibration (i.e. estimating the parameters $\omega_0$ and $\zeta$), one can leverage the analytical formulation for imputation directly. A notable advantage here is data efficiency, as the parameters can be estimated using much fewer samples. However, such simplified modelling can fail to capture complexities such as individual variability,  fatigue, and the impact of muscle dynamics. Standard differential equations are more suitable in low-data regimes and when the dynamics are well understood. }

\update{\textbf{Non-rigid Dynamics.} \textit{Neural differential equations}, on the other hand,  allows for more flexibility as it learns the dynamics function ($\mathbf{g}_\theta$) in a data-driven way while incorporating neural networks for advanced feature extraction. This approach is especially useful when the true dynamics are unknown or too complex to describe analytically, enabling its application to a wider range of sensory problems. Additionally, it is more powerful when the data is noisy \cite{Goyal2023,goyallearning}. However, the effectiveness of data-driven methods depends on the quantity and diversity of the data. With the growing availability of large sensory datasets, and given that our approach leverages only \textbf{unlabeled} data, we see significant potential for its application.  }

\subsection{NeuralPrefix Archiecture}
\label{sec:architecture}

Our architecture is a continuous-time encoder-decoder pipeline similar to \cite{rubanova2019latent} trained to reconstruct the masked data frames. During training, frames of the complete samples are randomly masked to imitate intermittency. The model is then trained to recover the masked samples by conditioning on the observed ones.

We follow the standard blueprint of  encoder-decoder approaches. We design continuous-time \update{enconder} and decoder. In the encoding \update{stage,} we perform \textit{forward consolidation} in which all the available frames are encoded into one (condensed) latent state vector. Similar to recurrent \update{approaches,} we process the frames sequentially. Unlike them, we base the latent state transition on NeuralODE.
For generation, we apply \textit{ backward unrolling}; an autoregressive process that continually generates all the missing \update{frames} (in reverse order) from the latent state vector. Directly predicting the target frame in ``one go'' is typically employed when operating in the in-domain (i.e., training and test samples are expected to be homogeneous).  However, this doesn't generalise to out-of-domain settings.  To resolve this, we integrate a \textit{modular frame generation} component. Roughly speaking, we think of the generated frame as mostly a perturbation of the last observed frame. Thus, the decoder is tasked with estimating the perturbations as dictated by the cross-frame dynamics rather than the appearance. Fig.~\ref{fig:arhcitecture} provides an overview of the NeuralPrefix architecture, with details of its components outlined below. 

\textbf{Encoder \textit{(forward consolidation)}.} Convolutional gated recurrent unit \texttt{ConvGRU} is used in the encoder $\mathcal{G}_{E}$ to capture the spatio-temporal patterns of the sensory data. \update{ConvGRU is an extension of the standard Gated Recurrent Unit (GRU) designed specifically to process spatio-temporal data, such as video frames or sequences of images. It combines the recurrent nature of GRUs, which handle temporal dependencies, with convolutional operations, which handle spatial patterns. Check \cite{ballas2015delving} for more details.} At each \update{step,} the input is processed using convolutional embedding prior to applying  \texttt{ConvGRU}. The latent state is continually estimated at all times as an ODE solution using Equation~\eqref{eq:ode_solve} where a fully connected network \texttt{MLP} approximates the dynamics component $\mathbf{g}_\theta$. As shown in the unrolled view in Figure~\ref{fig:arhcitecture},  which is passed to the decoder to be used for frames imputation.

\textbf{Decoder \textit{(backward unrolling).}} The decoder $\mathcal{G}_D$
generates the missing samples at the specified time steps. First, using ODESolve, we  estimate the latent codes $\big\{ \mathbf{h}(m_i) \big\}$ (green circles in Fig. \ref{fig:arhcitecture}) at the target temporal coordinates $\big\{ m_i \big\}$.   Then, the decoder iteratively maps the latent codes into the target frame $\mathbf{x}^\prime(m_i)$.

\textbf{Modular Frame Generation.} An important question that we address in NeuralPrefix is how to generate frames for unseen modalities? While it is possible to train the decoder to generate the whole frame directly, this approach creates issues in out-of-domain \update{settings}. Essentially, it will cause the appearance features to be ``baked'' into the generative model weights. Thus, the output will resemble the training data even when operating on \update{a new} modality. To address the issue, we predict the \textit{frame differences} components rather than the actual frames. 
Specifically, we adopt a modular generation in which we generate a set of elementary motion components that when combined with the last observed/generated frame recovers the new frame. 
We predict the motion flow information that represents cross-frames motion of pixels. \update{Then,} we compose the actual frame guided by the predicted motion flow and the last observed (i.e. seen or generated) frame.  To understand this, one can think of the generated frame as a weighted sum of a transformed (wrapped) version of the last frame and the new contents (appearance residuals) that will appear only in the new frame. Formally \cite{shen2024ladder, yu2022deep}, the generated frame is :

\begin{align}
    \label{eq:composition}
    \mathbf{x}^\prime(m_i) & = \Gamma_{m_i} \odot C_{m_i} + (1 - \Gamma_{m_i}) \odot R^\prime_{m_i} \\
    \label{eq:content}
    C_{m_i} & = \mathcal{W}(F_{m_i}, \mathbf{x}^\prime(m_{i-1}) )
\end{align}
where $\Gamma$ , $\mathcal{W}$  and $F$  denote a binary composition mask, the image warping operator and the forward motion flow; respectively. We task the decoder to learn the components of the equation above. The mask is ensured to be bounded in the range [0,1] by applying the sigmoid function. Notably, the motion flow is readily interpretable (akin to optical flow in the vision domain) and can offer insights into the system behaviour under various settings (Sec. \ref{sec:out_of_domain} ).   We can see that the content term is created as a warping of the frame generated by the model at the last timestep. during training, we use a residual loss $\mathcal{L}_{\text{residual}} = R^{\prime}_{m_i} - R_{m_i}$  where $R_{m_i}$ is calculated from the ground truth as the pixel-wise difference between consecutive frames. Additionally, we employ a content loss $\mathcal{L}_{\text{content}}$ as the MSE between all the frames generated at $m_i$ and the corresponding ground truth.

\begin{figure}[!t]
    \centering
    \includegraphics[width=1\linewidth]{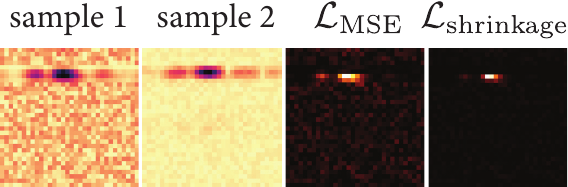}
    \caption{Comparison of pixel-wise losses. 
    The proposed $\mathcal{L}_{\text{shrinkage}}$ focuses on foreground regions and less on the background and noise as opposed to the uniform weighting by $\mathcal{L}_{\text{MSE}}$. }
    \label{fig:shrinkage}

\end{figure}

\textbf{Shrinkage Loss.} The sparsity of sensory (e.g. RF frames) results in an imbalance between the foreground (e.g. RF reflections from hand and body) and background pixels (e.g. pixels with low energy). 
Making the reconstruction tricky compared to natural images. For example, the  Mean Squared Error (MSE) 
 penalises the reconstruction error for the whole image pixels in a uniform way.
 While this is desirable for natural images with rich content and texture, it can be 
 sub-optimal for data where zero pixels and low energy noise dominate. 
 Without proper weight, the loss {\em unrealistically} small even for poor predictions. To mitigate the issue, we suggest attaching more importance to the few, but hard to reconstruct, informative pixels. Inspired by a similar issue in object detection \cite{lin2017focal} and object tracking \cite{lu2020deep}, we adopt the shrinkage loss $\mathcal{L}_{\text{shrinkage}}$ (Equation (6) in \cite{lu2020deep}) that down weights the easy pixels. Figure~\ref{fig:shrinkage} contrasts it to the MSE. Putting it all together,  the model is trained in end-to-end manner with the following loss:
\begin{equation}
    \mathcal{L}(\mathcal{G}) = \lambda_{1} \mathcal{L}_{\text{shrinkage}} +\lambda_{2} \mathcal{L}_{\text{residual}} +\lambda_{3} \mathcal{L}_{\text{content}}
\end{equation}
where $\lambda_{1}$, $\lambda_{2}$  and $\lambda_{3}$ are hyperparameters controlling the contribution of each loss term.

\section{Experimental Evaluation}


\subsection{Datasets}
\label{sec:eval_dataset}

We test NeuralPrefix on three publicly available sensory datasets. The first two datasets contain radar sensor measurements of human gestures, while the third is a pressure mat dataset capturing daily human activities such as locomotion, exercises, and resting. For all datasets, NeuralPrefix treats the samples as 4D heatmaps with time, spatial (width \& height), and channel dimensions. Our motivation for considering the radar datasets stems from their prominence in the RF domain \cite {wang2022placement} and the significant impact of deployment considerations on the measurements \cite{wang2022placement}. Even simple changes in either the \update{receivers'} placements, the user location, the user's orientation or the furniture in the environment will alter the received signal considerably. While this \update{holds} for other sensing modalities such as IMUs for human activity recognition, the domain shift is often more challenging in RF \cite{nirmal2021deep}. This presents a nice playground for testing zero-shot imputation performance under deployment domain shift (\textit{domain-OOD}).  Specifically, by training the model in one configuration (user orientation and device placement) and testing it on the rest, we can assess the system's performance in unseen domains. Additionally, the radar and the pressure mat datasets enable us to evaluate the system's performance on unseen modalities (modality-OOD) by training on one dataset and testing on the others. Details of the datasets are provided below.

\begin{itemize}
    \item \textit{Google's Soli Dataset} \cite{wang2016interacting}. This dataset contains 11 gestures from 10 subjects collected over multiple sessions. Each frame of the dataset is represented as a Range Doppler Image (RDI). In an RDI, one axis denotes the radial distance (or range) between the hand and the radar, while the other axis represents velocity. The pixel intensity corresponds to the energy reflected from objects (e.g. hands). The Soli sensor was employed in various applications, including gesture interaction with wearables \cite{wang2016interacting}, and objects \cite{vcopivc2019missing}, as well as material sensing \cite{vcopivc2022solids}. 
    
    \item \textit{MCD dataset} \cite{li2022towards}. This dataset contains RF measurements (using TI AWR1843 mmWave radar sensor) from 750 domains (6 environments $\times$ 25 subjects $\times$ 5 locations). The frames are Dynamic Range Angle Image (DRAI) measurements of subjects performing hand gestures.  In the RF datasets, we use half of the data for training and the other half for testing. 
    \item \textit{Intelligent Carpet dataset}\cite{luo2021intelligent}.  This dataset is used as an out-of-domain modality for the zero-shot experiment in Sec.~\ref{sec:out_of_domain}. A random subject data (\textit{24\_10\_TZ}) was used for testing.

\end{itemize}

\subsection{Metrics}

To report the performance, standard metrics used in spatio-temporal data imputation are employed. Namely, the Mean Squared Error (MSE), the Structural Similairy (SSIM) \cite{wang2004image}, Peak Signal-to-Noise Ratio (PSNR), and Learned Perceptual Image Patch Similarity (LPIPS) \cite{zhang2018unreasonable}.  These metrics capture \update{the} distance/similarity between the model's prediction and the ground truth. \update{SSIM is bounded in the range $[-1,1]$.} Additionally, we report the performance on a downstream task of hand tracking after applying the imputation.

\subsection{Baselines and Training}
\label{sec:eval_baseline_details}

In the evaluation, we compare against \update{several traditional approaches commonly used as efficient baselines in time series imputation works \cite{du2024tsi, ma2020transfer}. Additionally, we consider a theoretically-principled training-free approach based on optimisation and a spatio-temporal interpolation deep model. Specifically, we consider the following traditional approaches:}

\begin{itemize}
    \item \update{\textbf{Mean.} Missing values are imputed as the mean of the observed values.}
    \item \update{\textbf{Last Observation Carried Forward (LOCF).} Missing values are replaced with the last observed value for that variable.}
    \item \update{\textbf{Expectation Maximisation (EM) \cite{dempster1977maximum}}. We adopt the expectation-maximisation algorithm to impute the missing frames in $\mathbf{x}$. We consider imputing the temporal trajectory of each pixel $\mathbf{x}_{h,w,c} = \{x_{h,w,c}(o_1), \cdots, x_{h,w,c}(o_q)\}$ where $h,w$ denotes the spatial coordinates, $c$ is the channel coordinate and $o_1, \cdots , o_q$ is the set observed temporal coordinates. We assume Gaussian Mixture Model (GMM) as the data distribution. The algorithm alternates between probabilities computation (E Step) and parameters update (M Step). After convergence, imputation is done by sampling from the fitted GMM $\sim \sum_{k=1}^{K} \pi_k \mathcal{N}(x \mid \mu_k, \sigma_k^2)$ where $K = 3$ is the number of components. $\pi_k , \mu_k$ , and $\sigma_k^2$ are the weight, mean, and variance of the $k$-th Gaussian component; respectively.} 
    \item \update{\textbf{Optical Flow (OF) \cite{horn1981determining}.} OF leverages the vector field describing the apparent motion of each pixel between two adjacent
frames for imputation. }

\end{itemize}

\update{Additionally, we consider  the training-free baseline; \textbf{Optimal Transport (OT)} \cite{khamis2023earth}.} OT is a mathematical framework that enables us to compute the most efficient way to ``move'' one set of data points (such as probability distributions, images, or signals) to another while minimising some notion of cost. In the context of data interpolation, OT can be used to create smooth transitions between two frames by finding an optimal transportation plan between them. Specifically, we treat two frames as probability distributions $P$ and $Q$. Then, we compute the optimal transportation plan using Equation 2 in \cite{khamis2023earth}. \update{The imputed data is computed using the transportation plan as a series of intermediate distributions that gradually transition from $P$ to $Q$.} For the training-based baseline, we consider a generative model. For fair evaluation, we compare against a Neural ODE -based architecture \cite{kanaa2021simple}. 

\update{\textbf{Training and Implementation Details.} In all experiments, the model is trained for 500 epochs on a Nvidia RTXA6000 GPU. The training takes about 6 hours.  Adam optimiser is used, and the learning rate is set as ${10}^{-3}$, then exponentially decayed at a rate of 0.99 per epoch. We consider a batch size of 64. The parameters $\lambda_{1}$ , $\lambda_{2}$  and $\lambda_{3}$ are set to  0.05, 0.5 and 1; respectively.}

\begin{figure*}
    \centering
    \includegraphics[width = 1\linewidth] {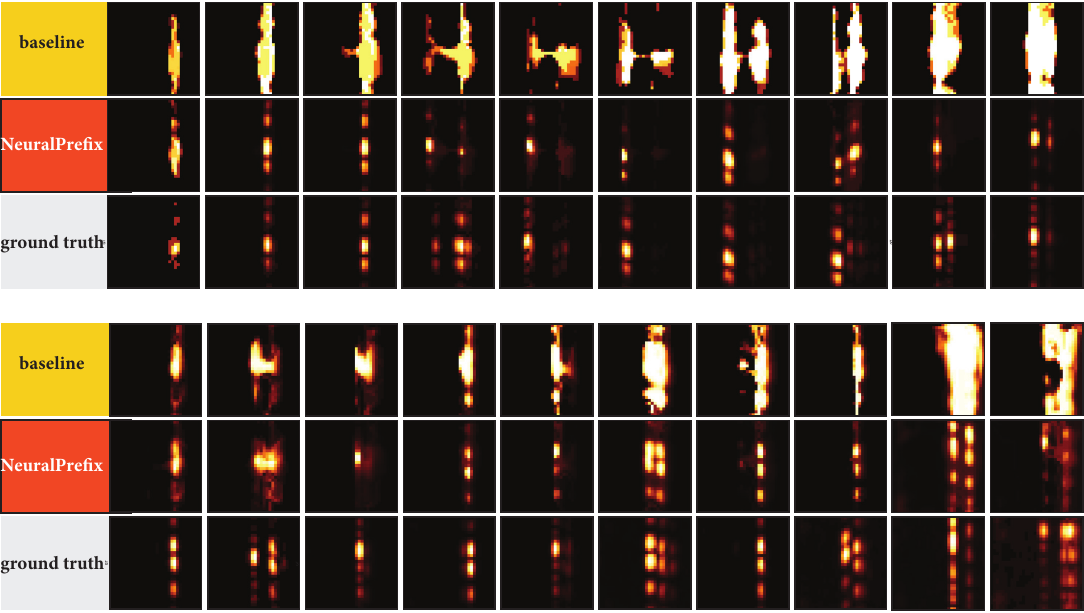}
    \caption{Qualitative comparison of intermittent data reconstruction using NeuralPrefix vs Neural ODE video generation baseline  \cite{kanaa2021simple} on MCD dataset. For visual convenience, we show only the dropped samples (i.e. w/o preceding and the following samples). It can be seen that NeuralPrefix significantly outperforms the baseline. }
    \label{fig:qualitative_comparison}
\end{figure*}

\subsection{Zero-shot Imputation Performance (domain-OOD)}
\label{sec:eval_imputation}

\begin{table}[h!]
    \centering
    \caption{Imputation Performance}
    \label{tab:imputation_performance}
    \begin{tabular}{llcccc}
        \toprule
        \textbf{Datasets} & \textbf{Model} & \multicolumn{4}{c}{\textbf{Interpolation}} \\ 
        \cmidrule(lr){3-6} 
        & & \textbf{SSIM}$\uparrow$ & \textbf{MSE}$\downarrow$ & \textbf{LPIPS}$\downarrow$ & \textbf{PSNR}$\uparrow$ \\ 
        \midrule
        \multirow{7}{*}{MCD} 
         & Mean & 0.7534  & 0.0017 & 0.0437 & 31.4275  \\
        & LOCF & 0.7512  & 0.0023 & 0.0506  &  30.5471 \\ 
       & EM & 0.6582  &  0.0035& 0.0952  & 28.9941  \\ 
         MCD& OF & 0.7831  & 0.0018 & 0.0370 &   34.4903\\ 
        
         & OT & 0.7143  &0.0041  & 0.1734 &  33.7536 \\ 

        & Baseline \cite{kanaa2021simple} & 0.8726  & 0.0015  & 0.1225 &  28.2929 \\ 
        & NeuralPrefix & 0.9399 & 0.0007 & 0.0752 & 31.8556  \\ 
        \midrule
        \multirow{7}{*}{Soli} 
        & Mean & 0.9561  & 0.0017 & 0.0441 &  32.2322 \\ 
        &  LOCF & 0.9646 & 0.0019  & 0.0331 & 32.1244 \\ 
        & EM & 0.9474  &  0.0026  & 0.0361 &  30.0411 \\ 
        Soli& OF & 0.9675  & 0.0017 & 0.0313 &  31.9493 \\ 
        
        & OT & 0.8937 & 0.0325 & 0.1815 & 24.0149  \\ 
        & Baseline \cite{kanaa2021simple} & 0.9497 & 0.0025 & 0.1524 & 26.1049  \\ 
        & NeuralPrefix & 0.9817 & 0.0006 & 0.0260 & 32.4366  \\ 
        \midrule
        \multicolumn{2}{c}{} & \multicolumn{4}{c}{\textbf{Extrapolation}} \\
        \cmidrule(lr){3-6} 
        & & \textbf{SSIM}$\uparrow$ & \textbf{MSE}$\downarrow$ & \textbf{LPIPS}$\downarrow$ & \textbf{PSNR}$\uparrow$ \\ 
        \midrule
         MCD & NeuralPrefix & 0.8569 & 0.0017 &  0.1524 & 27.4833  \\
         \midrule
         Soli & NeuralPrefix &0.9202 & 0.0076 & 0.1930 & 30.4366 \\ 
         \bottomrule
    \end{tabular}
\end{table}

We first evaluate the model's imputation performance in two modes; interpolation and extrapolation. In each mode, we drop 50\% of the data samples at the points ${m_i}$. In interpolation mode, the samples are taken at random position ${m_i}$  between $t_1$ and $t_N$. In extrapolation mode, the model observes the first half of the window and then reconstructs the remaining half. We set the window size to 10.

Table~ \ref{tab:imputation_performance} shows the imputation performance.  \update {Quantitatively, NeuralPrefix outperforms all the baselines on both Soli and MCD datasets for most metrics. The performance gap is bigger in the MCD dataset. One can notice that the traditional approaches Mean, LOCF and OF are slightly worse than the learned approaches on the Soli dataset.  However, the performance gap is much bigger in the MCD case. This is because MCD dynamics are more complex as the data contains reflections from multiple objects (subject's arm, subject's body and background reflections). On the other hand, Soli data is dominated by hand reflections with minimal background interference. Also, MCD has a much lower sampling rate than Soli (20 FPS in MCD vs 40 FPS in Soli). Thus, cross-frame transitions are much less smooth in MCD.} 
\update{The results reveal that Expectation Maximisation is the least performing on the MCD dataset despite being much more computationally demanding than simpler approaches such as Mean and LOCF.  EM has several limitations, including data modelling assumptions (e.g., assuming Gaussian Mixture Models). Recent works \cite{ma2021emflow, richardson2020mcflow} address this by combining EM with deep models. Optimal Transport (OT) also underperforms NeuralPrefix as it relies only on the physical principles (i.e., the most efficient transportation of pixels between frames) without leveraging the knowledge in the training dataset. We exclude traditional approaches from further evaluations.}

Qualitatively, it can be seen from Figure~\ref{fig:qualitative_comparison} that the learned baseline output is smeared out in regions with high energy. NeuralPrefix, on the other hand, preserves the sharp details as its frame composition (Equation \eqref{eq:composition} and Equation\eqref{eq:content}) explicitly learns the motion flow and the difference (residual) across frames.

\subsection{Impact on Downstream Task}
\label{sec:eval_downstream_task}

\begin{figure}
    \centering
    \includegraphics[width=1\linewidth]{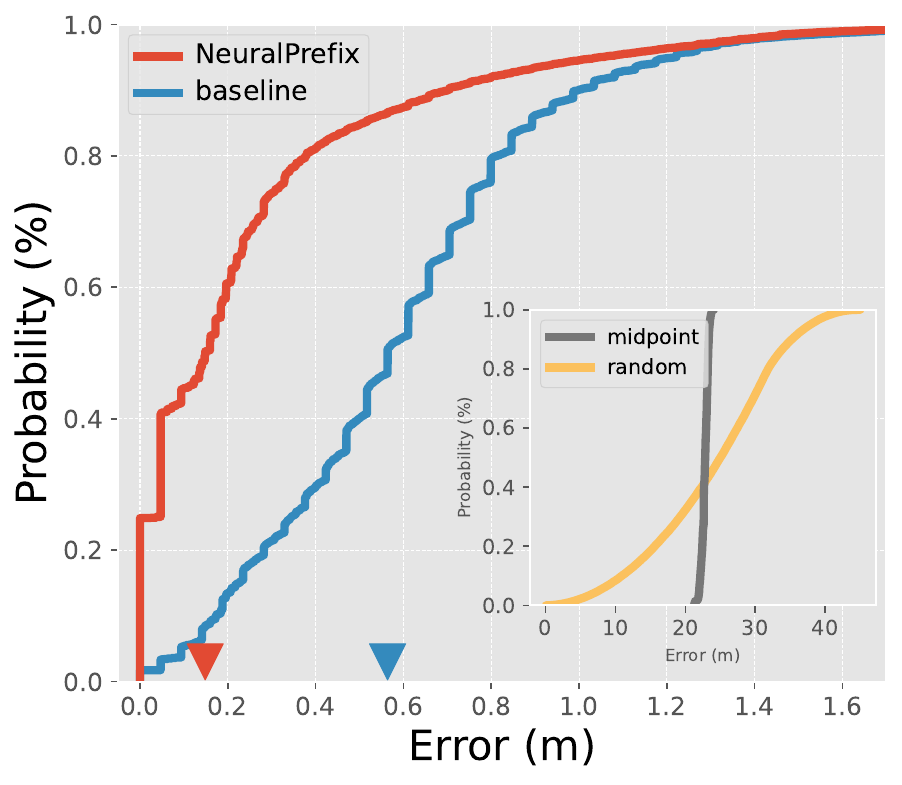}
    \caption{\textbf{Hand tracking through intermittent stream in MCD dataset.} CDF curves for the hand tracking error demonstrates the superiority of NeuralPrefix. Median error are denoted by triangular markers on x-axis. Inset plot shows the trivial baselines in which the hand position in the missed frames is set either randomly (`random') or as the median point of the range/angle bounds (`midpoints').  }
    \label{fig:CDF_tracking}
\end{figure}

While the results in Sec.~\ref{sec:eval_imputation} quantify the visual resemblance between the model's predictions and the ground truth, it is also informative to quantify the impact of this on a downstream task.  We perform hand tracking on the MCD frames generated by NeuralPrefix. Note that DRAI sequences encode the location and motion information of the hand and the body. We first locate the hand in each frame as the centre $(r,\theta)$ of the blob with the highest magnitude (following Sec. 3.2.7 in \cite{li2022towards}), where $r$ and $\theta$ denote the range and angle polar coordinates; respectively. We assume that the location in the ground truth frames is perfect (i.e. error = $0cm$) and calculate the deviation with respect to it.  The error is reported as euclidean distance $d = \sqrt{r_{i}^2 + r_{g}^2 - 2r_{i} r_{g} \cos(\theta_{i} - \theta_{g})}$, where the subscripts $i$ and $g$ denote the coordinates in the imputed and ground truth frames; respectively. The results are shown in Figure~\ref{fig:CDF_tracking}.
NeuralPrefix median error is $14.85cm$ compared to $56.85cm$ by the baseline. This \update{is a} $3.8$x improvement in the tracking accuracy.  This is a direct consequence of the improved visual quality of NeuralPrefix's output. As the details of frames are sharper in NeuralPrefix case, the hand's blob can be located accurately.

\subsection{Zero-shot Imputation Performance (modality-OOD) }
\label{sec:out_of_domain}

This section evaluates the zero-shot performance of NeuralPrefix. The model is trained on one dataset (seen) and tested on another (unseen) dataset from a different sensory domain. This is done without re-training or adaptation. Thus, testing the model's zero-shot generalisation capacity. The results in  Table. ~\ref{tab:zero_shot_performance} are promising, demonstrating good performance on the unseen dataset with only minor degradation compared to the in-domain reference (where the target and source datasets are the same). Note that although MCD and Soli are both RF datasets, they were collected using different sensors with different configurations, capturing semantically different measurements. The findings become even more interesting in the second part of Table. ~\ref{tab:zero_shot_performance}, where we test on the Intelligent Carpet dataset. In this case, Soli is picked as the seen dataset, as it is visually more similar to the carpet dataset than the MCD.

\begin{table}[h!]
    \centering
    \caption{Zero-shot Data Imputation Performance}
    \label{tab:zero_shot_performance}
    \label{table:comparison}
    \begin{tabular}{llcccc}
        \toprule
        \textbf{Seen} &  \textbf{Unseen} &  \multicolumn{4}{c}{\textbf{ performance}} \\
        \cmidrule(lr){3-6} 
        & &  \textbf{MSE$\downarrow$} & 
        \textbf{SSIM$\uparrow$} &
        \textbf{LPIPS$\downarrow$} & 
        \textbf{PSNR$\uparrow$ }  
        \\
        \midrule
        MCD 
        & Soli & 0.0012 & 0.9402 & 0.0843 & 29.1026 \\
        \multicolumn{2}{c}{in-domain reference} & 0.0006 & 0.9817& 0.0260 & 32.4366 \\
        \midrule
        Soli
        & MCD&0.0012 & 0.8888 & 0.1269 & 29.1438 \\
       \multicolumn{2}{c}{in-domain reference} & 0.0007 & 0.9399 & 0.0752 & 31.8556 \\
       \bottomrule
       Soli & Carpet & 0.0016 & 0.8820  & 0.1335 & 27.8321 \\
        \bottomrule
    \end{tabular}
\end{table}

\begin{figure}
    \centering
    \includegraphics[width= 1\linewidth]
    {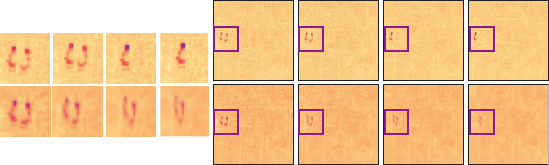}
    \caption{\textbf{Zero-shot Imputation} on Intelligent \update{carpet} dataset using NeuralPrefix trained on Soli dataset. (Top) ground truth (bottom) data imputed by the model.}
    \label{fig:out_of_domain_carpet}
\end{figure}

Upon further analysis, we observe the model can still predict the dynamics on the unseen dataset correctly. In Figure~\ref{fig:out_of_domain_carpet} (zoomed-in part), the ground truth reveals the subject motion transition from two feet on the mat to just one. Visually, NeuralPrefix wasn't able to recover the frames exactly. Yet, the forward flow predicted by the model ($F_{m_{i}}$ in Eq.\eqref{eq:composition}) caused the active blob to gradually shrink closing the gap between the two shapes.  

\begin{figure}[t!]
    \centering
    \includegraphics[width=1\linewidth]{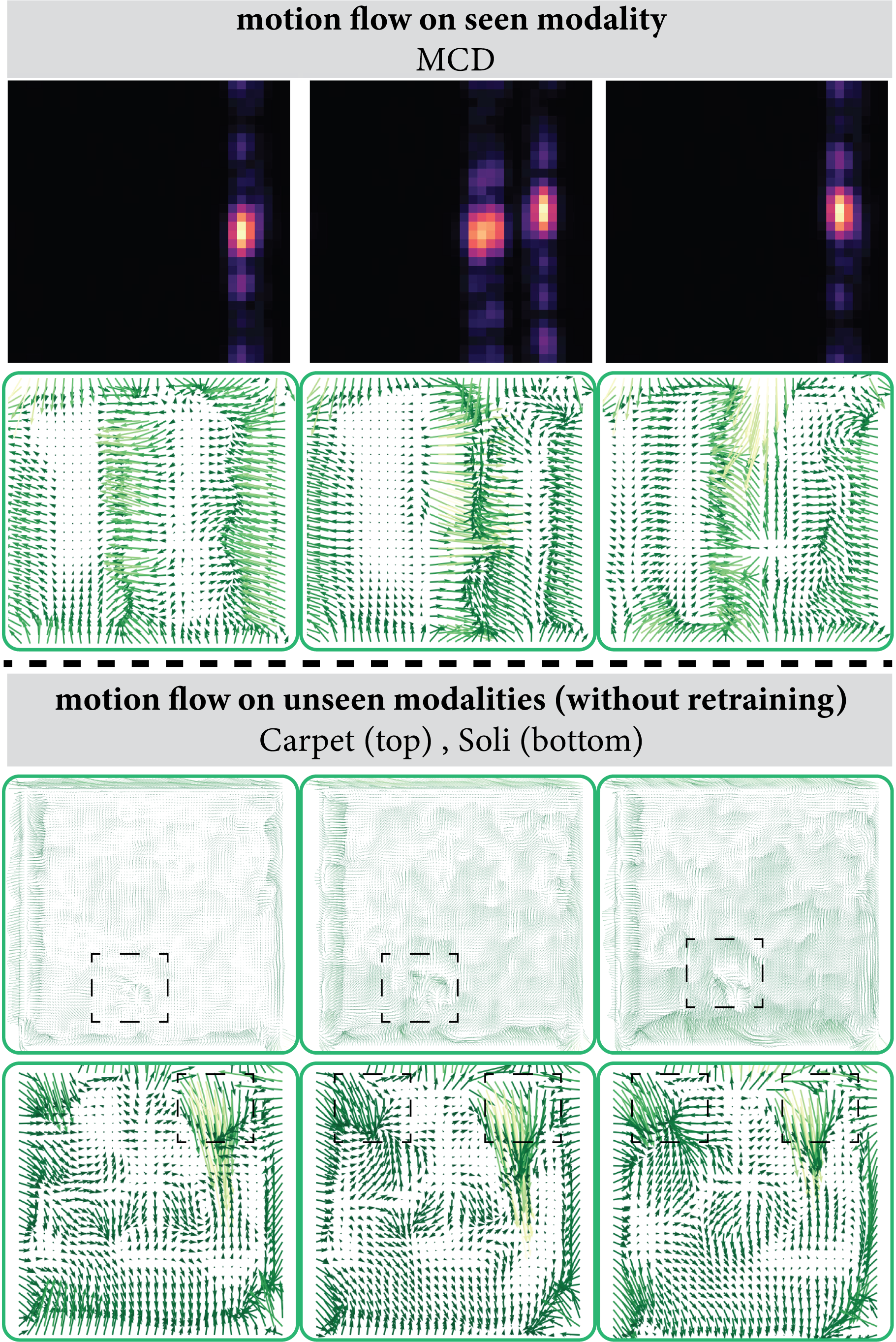}
    \caption{\textbf{NeuralPrefix learned Motion Flow $F_m$ (Equation \ref{eq:composition})  generalises to unseen modalities. }}
    \label{fig:motion_flow}
\end{figure}

Recall from Sec. \ref{sec:architecture} that the motion flow information ($F_{m_i}$ in Equation. \ref{eq:composition}) is one of the outputs of the modular frame generation component. Since the motion flow can be visualised as vector flow, we can understand NeuralPrefix's behaviour in unseen modalities by inspecting them. Figure \ref{fig:motion_flow} shows the motion flow in seen and unseen modalities. In this case, NeuralPrefix was trained on MCD dataset and tested on Soli then on Carpet without fine-tuning or re-training. First, we note that the motion flow correctly captures the pixels motion across frames in MCD dataset. Interestingly, we noticed that occasional inaccuracies in motion vectors positions can happen without impacting the final frame quality! This happens because the final frame is a composition of the motion flow $F_{m_i}$ and residual $R_{m_i}$. We notice that when the network fails to exactly recover the motion flow, the estimated residual $R_{m_i}$ compensates for it. 

Moving to the unseen modalities, the dynamics produced for the unseen modalities (Carpet and Soli) are significantly different from the reference modality (MCD) as shown in Figure \ref{fig:motion_flow}. For example, on Carpet,  the vectors magnitudes are much smaller than that of either Soli or MCD. Signifying the true fact that most pixels in the frames belong to the background. This suggests that NeuralPrefix learned transferable dynamics from MCD and didn't memorise to the seen dynamics. The dashed box superimposed on the motion flow frames indicates the position of the active object (human foot in Carpet, reflection from hand in Soli) in the original frames. Ideally, these regions should contain the highest magnitude of motion. While it is observable in Soli, it is less observable in Carpet due to the high noise in the background.

\subsection{Computationally Adaptive NeuralPrefix}
\label{sec:comp_adaptive}

\update{One interesting aspect of our architecture is its computational efficiency and elasticity. During the inference stage, NeuralPrefix takes an average of 124 milliseconds for processing a single sequence $\mathbf{X}$ when the window size $t$ is set to 10. For a window size of 20, it takes an average of 196 milliseconds. Additionally, NeuralPrefix offers a very natural way to trade off the accuracy and computational load without changing the current architecture. Thus, having the potential to adapt to the computational requirements of various ubiquitous devices.} Specifically, the ODESolve in NeuralPrefix can be expedited by limiting the number of iterations. This can be done very simply by changing the tolerance parameter which controls the accuracy of the ODE solution. It sets the acceptable error threshold that the numerical solver uses when approximating the solution at each time step.

We experimented with the tolerance of the ODESolve in the Encoder and Decoder components. The default tolerance is set to $10^{-5}$. On the MCD dataset, when increasing it to $10^{-3}$,  we gain a processing speedup by 20\% without impacting the quality of the frames (SSIM:0.9122 compared to 0.9399 in the default case). A major speedup of 50\% can be gained by setting the tolerance to 0.5. \update{However,} it comes at the cost of degraded quality of output (SSIM: 0.8134). Note that, this approach is more flexible than weight pruning or quantisation \cite {cheng2024survey}, since the tolerance can be adjusted even after the model has been deployed.

\section{Discussion and Limitations}
\label{sec:dicussion}

In our effort to demonstrate the feasibility of \update{zero-shot} imputation through the development of NeuralPrefix, we learned a number of lessons. We highlight both limitations and opportunities that can inspire future research in this field below. 

\textbf{Performance on Synthetic Measurements.} In our initial investigation, we considered the Widar 3.0 dataset \cite{zheng2019zero}. However, its performance was lower than that of the datasets used in this paper. Visually,  the cross-frame apparent dynamics of the Widar 3.0 samples are less smooth, with many abrupt changes. This is because the frames are analytically synthesised (from multiple receivers) to approximate a body velocity profile (BVP) using compressive sensing techniques. This suggests that heavily pre-processed (or analytically synthesised) data are harder to learn from. Our findings suggest that heavily pre-processed or analytically synthesised data are more challenging to learn from. Therefore, we selected the Soli, MCD, and Carpet datasets instead. 

\textbf{Prior Integration.}  We noticed that the motion flow \update{vectors magnitudes} can be exaggerated for unseen modalities.  The current pipeline offers opportunities for integrating prior beliefs. One of which is regularising the motion vector. For example, in the human sensing domain, \update{human} motion typically has a known range. This can be used analytically to estimate the maximum magnitude \update{of the} motion vectors and regularising against it. Achieving this without re-training the whole model can be done by integrating adapter techniques \cite {kang2024sf} that are employed in unsupervised Test Time Adaptation (TTA).

\update{\textbf{Robusteness Evaluation.} While our method shows promising OOD results on the considered setup, further evaluation is needed to understand its behaviour and limitations in more challenging scenarios. For example, long missingness windows,  edge cases not represented in the training dataset, and novel missingness patterns \cite{qian2024unveiling}.}

\section{Conclusion}

This paper introduces NeuralPrefix, the first attempt to formalise and address zero-shot spatio-temporal sensory data imputation. Our evaluation demonstrates its practical feasibility and effectiveness on real datasets. NeuralPrefix is carefully designed to ensure flexibility of the imputation mode and generalisation in unseen contexts.  This capability enables various applications in the pervasive sensing domain such as modality compensation and early action prediction. The current work can be extended in several ways. For example, quantifying the latency of extrapolation for early action prediction and speeding up the model by optimising the forward evaluation of the ODE solver.

\bibliographystyle{acm}
\bibliography{refs}

\end{document}